\documentclass{article} % For LaTeX2e
\usepackage{nips15submit_e,times}
\usepackage{hyperref}
\usepackage{url}

\usepackage{color, amssymb, amsmath}
\usepackage{amsmath, amssymb, amstext, amsthm, color, varioref}

\theoremstyle{definition}

\theoremstyle{definition}

\theoremstyle{plain}

\theoremstyle{plain}

\theoremstyle{plain}

\theoremstyle{plain}

\theoremstyle{plain}

\theoremstyle{plain}

\title{Clustering is Easy When $\ldots$ What?}

\author{
Shai Ben-David \\
Cheriton School of Computer Science\\
University of Waterloo \\
Waterloo, Canada\\
\texttt{shai@uwaterloo.ca}}
\nipsfinalcopy % Uncomment for camera-ready version

\begin{document}

\maketitle

\begin{abstract}

It is well known that most of the common clustering objectives are NP-hard to optimize. In practice, however, clustering is being routinely carried out. 
One approach for providing theoretical understanding of this seeming discrepancy is to come up with 
notions of clusterability that distinguish realistically interesting input data from worst-case data sets. The hope is that  there will be 
clustering algorithms that are provably efficient on such ``clusterable" instances. This paper addresses the thesis that the computational hardness of clustering tasks goes away for inputs that one really cares about. In other words, that ``Clustering is difficult only when it does not matter"\footnote{This phrase is in fact a title of a recent paper --  \cite{DLS}.} (the \emph{CDNM thesis} for short).
 
%While I believe that to some extent this may indeed be the case, I wish to point out that current results are still far from supporting this belief.
%This paper provides a survey of recent papers along this line of research and a critical evaluation their results. Our bottom line conclusion is that  that CDNM thesis is still far from being formally substantiated.

 I wish to present a a critical bird's eye overview of the results published on this issue so far and to call attention to the gap between available and desirable results on this issue. A longer, more detailed version of this note is available as \cite{ben2015computational}.
 
I start by discussing which requirements should be met in order to provide formal support to the the CDNM thesis. I then examine existing results in view of these requirements and list some significant unsolved research challenges in that direction.

\end{abstract}

\section{Introduction}
%The goal of this note is two-fold. First, I would like to provide a personally biased overview of the research concerning the computational complexity of clustering under data niceness assumptions.  A secondary aim of this paper is to call the attention to some such gaps in our current understanding of the issue of "easier data" and encourage further work along those directions that might  have otherwise seemed resolved.
%
%
%
%\subsection{Alternatives to worst-case for measuring computational complexity}
Computational complexity theory aims to provide tools for the
quantification and analysis of the computational resources needed
for algorithms to perform computational tasks. Worst-case complexity
is by far the best known, most researched and best
understood approach to computational complexity theory.  In particular, NP-hardness is a worst-case-instance notion. By saying that a task is NP-hard (and assuming $P \neq NP$), we imply that for every algorithm, there exist infinitely many instances on which it will have to work hard. However, for many problems 
this measure is unrealistically pessimistic compared to the experience of solving them for practical instances. A problem may be NP--hard and still have algorithms that solve it efficiently for any instance that is likely to occur in practice or any instance for which one cares to find an optimal solution for.

Here, we focus on clustering tasks that are defined as discrete optimization problems. Most of those optimization problems are NP-hard.
We wish to examine whether this hardness remains an issue when we restrict our attention to ``clusterable data" - data for which a meaningful clustering exists (one can argue that when there is no cluster structure in a given data set, there is no point in applying a clustering algorithm to it). In other words, we wish to evaluate to what extent current theoretical work supports the ``Clustering is difficult only when it does not matter" (CDNM) thesis. In this note, we provide a relatively high level view of some of the major relevant results. A more detailed version of our results can be found in \cite{ben2015computational}.
For the sake of concreteness, we will focus on two popular clustering objectives, $k$-means and $k$-median.

%\subsection{Outline of the paper}
We start this note by listing, in Section \ref{requirements},  what we think are requirements from notions of clusterability aiming to substantiate the CDNM thesis. In Section \ref{notions_clust}, we list various notions of clusterability that have been proposed in the context of this line of research. 
%%
%%These include: 
%%\emph{Additive perturbation robustness} (APR), \cite{AckermanB09}; 
%%\emph{Multiplicative perturbation robustness} (MPR), \cite{BiluL10};
%%$(\alpha, \epsilon)$ \emph{Perturbation Resilienc}e, \cite{BalcanL12};
%%\emph{$\epsilon$ -Separatedness}, \cite{OstrovskyRSS12};
%%\emph{Uniqueness of optimum},  \cite{BalcanBG09} (they call it $(c, \epsilon)$\emph{-approximation-stablility});
%%\emph{$\alpha$-center stability}, \cite{AwasthiBS12};
%%and \emph{$(1+\alpha)$ Weak Deletion Stability}, \cite{AwasthiBS10}.
%%

%The main body of this paper is an examination, in Section \ref{meet_req}, of how well do the current notions and results meet the requirements (of Section \ref{requirements}).  To get a sense of how strict a clusterability condition is, we consider an optimal clustering of data sets that satisfy that condition and examine the implied bounds on the ratio between the average distance of a data point to its own cluster center and the distance between centers of different clusters (or the distance of a point from centers of clusters it does not belong to). 

We then examine the results pertaining to the proposed notions of clusterability listed above, from the perspective of those requirements. Due to the conciseness of this note, we list here only some representative of our results and refer to the full version \cite{ben2015computational} for a more complete list.

Our conclusion is that the currently available theory is still far from substantiating the CDNM thesis. In particular, while additive perturbation robustness, with any non-zero robustness parameter, gives rise to algorithms that find the optimal clusterings in time polynomial in the input size and its dimension, as far as currently published results go, none of the proposed clusterbility condotions allows finding optimal clustering solutions in time polynomial in the number of target clusters, $k$, unless the corresponding parameters are set to values that hold only for extremely well clusterable data sets

%\footnote{The above consequences 
%of the required clusterability conditions 
%are obtained by examining the parameter values and constants that are implicit in the asymptotic formulation of the 
%efficiency results in the above cited papers. One should note that these negative statements reflect only the current state of knowledge, and are not proven lower bounds.
%For some of the above notions of clusterability, we also discuss lower bounds on the parameter values required to overcome the NP-hardness of the clustering tasks.}.
In Section \ref{conclusions} we summarize these discouraging results and highlight some implied open problems and propose directions in which this line of research should, in our opinion, proceed.

\section{Requirements from notions of clusterability} \label{requirements}
We begin by  stating requirements that (we believe) a notion of clusterability should satisfy to be applied for supporting the ``Clustering is Difficult only when it does Not Matter" (CDNM, in short) thesis. Those requirements are stated as qualitative, high level, statements. 
%We discuss more concrete quantitative formulations in Section \ref{meet_req} .

%\begin{enumerate} 

1. \emph{ It should be reasonable to assume that most (or at least a significant proportion of) the inputs one may care to cluster in practice satisfy the clusterability notion.}\\

% Some disclaimer is in place here;
 Of course, we do not have any way to guarantee that unseen practical instances will satisfy any non-trivial requirement. 
However,  this type of consideration can serve as a way to filter out clusterability conditions that are too restrictive. Furthermore, when a good data generative model is available, one can formalize requirements pertaining to a high probably of having the generated instances satisfy the given clusterability notion. 
 
2.  \emph{ In order to support the CDNM thesis, a notion of clusterability should be such  that there exist efficient algorithms that are guaranteed to find a good clustering (minimizing the objective function, or getting very close to it) for any 
input that satisfies that clusterability requirement}.

%\end{enumerate} 
The next two requirements may be more debatable. They are motivated by considering practical aspects of clustering applications.
Assume we do have some clusterability condition and a guarantee that the algorithm we are about to run is efficient on instances satisfying it. When we get some real input, there is no guarantee that it satisfies that clusterability condition.
% If it does not, and we run our algorithm, it may either run for too long or terminate with some sub-optimal solution. However, 
Since for most of the NP-hard clustering problems, there is no efficient way of measuring how far from optimal  a given clustering solution is, one may not being able to protect against bad solutions. 
%This consideration implies
%a third desirable requirement  -- the ability to distinguish between clusterable and non-clusterable input data sets. Namely,

3. \emph{ There exists an efficient algorithm for testing clusterability. Namely, given an instance $(X,d)$, the algorithm determines whether it satisfies the clusterability requirement or not.}

%Another advantage of having a notion of clusterability satisfy this requirement is that it will allow a direct evaluation of the extent to which the notion satisfies Requirement 1 above. Namely, having an efficient clusterability -checking algorithm, one could apply it to collections of representative practical clustering inputs from various domain and evaluate to what extent the clusterability requirement actually holds for such clustering tasks.\\

A forth, somewhat orthogonal, desiderata relates to existing common clustering algorithms. Namely, 

4. \emph{ Some commonly used clustering algorithm can be guaranteed to perform well (i.e., run in polytime and find close-to-optimal solutions) on all instances satisfying the clusterability assumption.}

Requirement 4 is important if our goal is to \emph{understand} what is happening nowadays in clustering work by providing a theoretical explanation for the success of common clustering algorithms on real data. However, even when failing it, requirement 2 may lead to the development of new clustering algorithms, which may have independent merits.

\noindent {\bf  The main Open Question:} \emph{Find a notion of clusterability that satisfies the requirements above (or even just the first two).}

\section{Notions of clusterability} \label{notions_clust}
In the past few years there have been several interesting publications along the lines described above,
showing that for various notions of clusterability there are indeed algorithms that find optimal clusterings in polytime for all appropriately clusterable instances. 
Below is a (possibly not exhaustive) list of major notions of clusterability  that have been discussed in that context\footnote{be ware that different papers use different terminology for similar notions (and similar terminology for different notions), so my choice of terminology below is not always consistent with other publications.}.  Most of these definitions can be applied to any center-based 
clustering objective. 
Due to space constrains, this version omits some of the technical details of the following definitions.
\begin{enumerate}
\item \textbf{Perturbation Robustness:} An input data set is perturbation robust if small perturbations of it do not result in a change of the optimal clustering for that set.
\begin{enumerate}
\item Additive perturbation robustness (APR) \cite{AckermanB09}\footnote{The definition of robustness, as well as the implied efficiency of clustering result, in \cite{AckermanB09} are particular cases of a more general definition and more general results of \cite{Ben-David06}}: An input set $(X,d)$ is $\epsilon$-APR if some optimal $k$-clustering $C$ remains optimal for any small (additive) perturbation of this input \footnote{Since this additive condition is not scale invariant, we implicitly add the assumption that the diameter of the input set, $\max_{x,y \in X}d(x,y)$, is at most 1 (otherwise the stability parameter should be multiplied by that diameter).}.

\item Multiplicative perturbation robustness (MPR) \cite{BiluL10}: An input set $(X,d)$ is $\alpha$-MPR if some optimal $k$-clustering $C$ remains optimal for any small (multiplicative) perturbation of this input.

%\item \cite{BalcanL12} propose the following relaxation of the MPR requirement:
%A data set $(X, d)$ is $(\alpha, \epsilon)$-\emph{perturbation resilient } if there exists some optimal $k$-clustering $C$  such that for every $d'$, if $ ~d(x,y) \leq d'(x,y) \leq \alpha d(x,y)$ for every $x, y \in X$, then for some $C' \in C_{\O} (X,d')$, $\mathcal{D}_{err}(C, C') \leq \epsilon$.
\end{enumerate}
\item \textbf{Significant loss of the objective when reducing the number of clusters:}
\begin{enumerate} 
\item \textbf{$\epsilon$ -Separatedness:} \cite{OstrovskyRSS12} discuss clustering w.r.t. the $k$-means objective. They define 
An input data set $(X,d)$ is $\epsilon$-\emph{separated for $k$} if 
the $k$-means cost of the optimal $k$-clustering of $(X,d)$ is less then $\epsilon^2$ times the cost of the optimal $(k-1)$-clustering of $(X,d)$.

\item \textbf{Weak Deletion Stability:} \cite{AwasthiBS10} An instance for $k$-clustering  satisfies the \emph{$(1+\alpha)$ Weak Deletion Stability} condition if, for its optimal clustering, removing any center $c_i$ and assigning all the points in its cluster to a different center $c_j$, results in an increase of cost of the clustering by a factor $\geq (1+\alpha)$. 
\end{enumerate}

\item \textbf{Center stability}: \cite{AwasthiBS12} An instance $(X,d)$ is \emph{$\alpha$-center stable } (with respect to some center based clustering objective $\O$) if for any optimal clustering of it, all points are closer by a factor $\alpha$ to their own cluster center than to any other cluster center.

\item \textbf{Uniqueness of optimum}: \cite{BalcanBG13}
%\footnote{This is a journal version of \cite{BalcanBG09}, where the definition and the main results were initially introduced.} define 
A data set is $(c, \epsilon)$\emph{-approximation-stable} with respect to some \emph{target clustering} $C_T$ if every clustering $C$ of $X$
whose objective cost over $(X,d)$ is within a factor $c$ of the objective cost of $C_T$ (on $(X,d)$) is $\epsilon$-close to $C_T$ (w.r.t. some natural notion of between-clustering distance). This condition rules out the possibility of having two significantly different close-to-optimal-cost solutions.
\end{enumerate}

\section{To what extent do the notions meet the requirements listed above?} \label{meet_req}
As varied as the above list of proposed notions may sound, it turns out that almost all (except for the additive perturbation robustness, which is also the only one that does not yield efficiency for large $k$) imply that data satisfying them is structured such that the vast majority of the data points can be assigned to compact clusters that are very widely separated (or that all but a small fraction of the clusters are such). We provide quantitative versions of this claim in Section \ref{pos_comput}. In fact, this common characteristic  of the notions is the main feature that is being used in showing that, under such conditions, clustering can be carried out efficiently.
While all of the above notions sound intuitively plausible (concrete arguments supporting that plausibility can be found in the papers presenting them), the quantitative values
of the clusterability assumptions are essential for evaluating that plausibility. We show (see \cite{ben2015computational}) that the currently known results concerning these notions yield the desired efficiency of computation only when the clusterability parameters are set to values that are beyond what one might expect practical inputs to satisfy.

% However, since 
%the actual values of the 
%parameters (that define the clusterability notions) determine both the runtime of the algorithms and the restrictiveness of the clsuterability conditions, these concrete values are needed when we wish to evaluate and the gap between what we currently know and the optimistic CDNM thesis.

\subsection{Computational efficiency vs realistic soundness of clusterable inputs} \label{pos_comput}

%An important distinction in this context concerns the meaning of hardness of computation. Clustering tasks where the clusters are determined by selecting cluster centers from the input set can clearly always be solved in time $m^k$ (where $m$ is the input size and $k$ is the number of clusters), by performing an exhaustive search over all possible cluster centers. For such problems, the term "feasible" usually refers to run time bounded by a polynomial in both $m$ and $k$. 
%%On the other hand, tasks like $k$-means, where the input set resides in some euclidean space, $\reals^n$, are often NP hard already for fixed values of $k$ (e.g., $k=2$) when the space dimension $n$ is a parameter of the runtime. For such problems, algorithms that have polynomial dependence on $m$ and $n$ may be considered ``feasible" even if they have exponential dependence on $k$. Of course, in order to have solutions that are also polynomial in $k$, the requirements on the input instances are more demanding. 

For all the above mentioned clusterability condition, it has been shown that when the clusterability parameters are set to sufficiently restrictive values, data satisfying those requirements allows polytime discovery of its optimal clustering solutions. However, examining the implications of those parameter settings to the input data satisfying them, we conclude that efficiency is obtained only for rather unrealistic data setup.  Typical examples of our results are (for a full list of those results see \cite{ben2015computational}):
\begin{itemize}
\item The values of $\epsilon$ for which $\epsilon$ -Separatedness is shown (in \cite{OstrovskyRSS12}) to allow $poly(k)$ clustering algorithms imply that, in the optimal clustering,  the average distance of a point from its cluster center
should be smaller than the minimal distance between distinct cluster centers by a factor of at least 200.
\item The values of parameters for which $(c, \epsilon)$ approximation stability  is shown (in \cite{BalcanBG09}) to allow $poly(k)$ clustering algorithms imply that, in the optimal clustering, for all but an $\epsilon$-fraction of the input points, the distance of a point to its own cluster center is smaller than its distance to the next closest center by at least 20 times the average point-to-its-cluster-center-distance.

\item The values of $\alpha$ for which $(1+\alpha)$ weak deletion stability is shown (in \cite{AwasthiBS10}) to allow $poly(k)$ clustering algorithms imply that, in the optimal clustering, the vast majority of the clusters are so distant from the rest of the data points that any point outside such a cluster is further from the center of that cluster by at least $\log(k)$ times the "average radius" of its own cluster.

\end{itemize}

\subsection{Efficient testability of the clusterability conditions}
When it comes to testing whether a given clustering instance satisfies any of the above clusterability conditions, a key point to note is that they are all phrased in terms of condition pertaining to the optimal clustering of the given data. Finding such optimal clusterings is NP-hard. Furthermore, there exist no efficient algorithm for testing, given a data set $(X,d)$ and a $k$ clustering of it, $C$, whether $C$ is an optimal clustering for $(X,d)$.

\subsection{Implications for common practical clustering algorithms}
Among all the works surveyed in this note, only one, the results of \cite{OstrovskyRSS12}, address (a feasible variant of) a practical algorithm - the popular Lloyd clustering algorithm.  It would be very interesting to come up with results showing that  some popular clustering algorithm (or an application of a practical approximation algorithm) efficiently yield  guaranteed good quality clusterings, under some other, or more relaxed, niceness of data conditions. 

The recent work of  \cite{AwasthiBC14}
can be viewed as a step in that direction. They ask under which separation condition do various convex relaxations exactly recover the ``correct" clustering.
However, that work addresses a different version of clustering problems, in which one assumes that the data is generated by some parameterized generative model (a balanced mixture of spherical Gaussians, in the case of that paper), and aims to recover those parameters.

\section{Conclusions} \label{conclusions} 
%Several notions of clusterability have been proposed so far. Depending on the values of the parameters defining those notions, each of them ranges from being very lenient to a  highly constraining data requirement.
%For each notion there is a parameter range so that, for data conforming to the clusterability requirement in that range, an optimal clustering can be rather trivially found. 
For each notion of ``easy clustering inputs" proposed so far, the parameter values that suffice for the currently available efficient clustering results turns out to be too strong requirements from  the practical significance perspective. 
The current failure to support the CDNM thesis may stem from various sources. First, of course, maybe the thesis is just false. My personal belief is that, while it may very well be the case that some practical clustering tasks are indeed computationally hard for some real data instances, there are many more cases where data of practical interest does yield not-too-hard-to-find meaningful clusterings (though, of course, most of the time we have no way of knowing whether those are optimal clusterings in any formal sense of optimality).

Another explanation to the shortcomings of current results is that they may just be an artifact of the algorithms and proof techniques that we currently have. 
%Maybe one could eventually come up with efficient algorithms that will cluster well under much less restrictive parameter settings of the clusterability notions listed in this note. Indeed, for most of those notions we do not have any close-to-matching computational hardness lower bounds. 
I doubt if that is indeed the case. 
%As mentioned above, for the notion of $\alpha$-center stability, the gap between the parameter values sufficient for efficient clustering and those that imply NP hardness is very small, $(1+\sqrt{2})$ vs $2$.
In fact, \cite{ShalevReyzin14}\footnote{This is a journal version of  \cite{Reyzin12} where the result initially appeared.} 
shows an NP-hardness result for center stability that almost matches the parameter values known to imply feasibility under such a condition.
%for any $\epsilon >0$ the problem of finding the optimal $k$-median clustering for $(2-\epsilon)$-center stable inputs is NP-hard.
%Furthermore, the $\alpha$-center stability is a central notions, in the sense that almost any other of the notions of clusterability discussed above implies it (or some variants of that condition), and the current results rely on those implications for proving the efficiency of clustering under those conditions.

I believe that part of the answer is that we have not yet discovered the appropriate notions of clusterability. The results surveyed in this paper indicate that notions of clusterability that aim to substantiate the CDNM statement should not be just a way of formalizing large between-clusters separation.
Apparently,  as demonstrated above, such assumptions become too restrictive before they yield efficient clustering results. 

\subsection{Call for a change of perspective on the complexity of clustering}
Finally, I would like to argue that if we really wish to model clustering as it is required and used in applications, the formulation of clustering tasks as computational problems should be revisited and revised.

All the papers surveyed above, as well as most of the current theoretical work on the computational complexity of clustering, focus on \emph{concrete clustering objectives} aiming to find the best clustering for a \emph{given number of clusters}.
However, the practice of clustering is widely varied. There are applications, like clustering for detecting record
duplications in data bases (say, records of patients from various hospitals and clinics), where the user does not set the number of clusters in advance, and aims to detect sets of mutually similar items to the extent that such sets occur in the input data. 
In other applications, like vector quantization for signal transmission or facility location tasks, while the objective function is usually fixed (say, $k$-means), 
 there is no implicit ``target clustering" and the usefulness of a resulting clustering is not diminished by having various different close-to-optimal solutions. 
 In some such applications $k$ is externally determined,  however, it is also common to consider optimizing some ``compression vs distortion" tradeoffs, rather than aiming for a fixed number of clusters. 

Furthermore, while the restriction of the problem of finding a good clustering to a given number of clusters $k$ may make practical sense when $k$ is small, for data sets that yield a very large number of clusters it is harder to imagine realistic situations in which that number, $k$, should be fixed independently of the particular input data set.
Still, most of the work surveyed above focuses on analyzing the asymptotic, w.r.t. $k$, computational complexity of $k$-clustering where $k$ is determined as part of the problem input. 

In many cases, the actual goal of clustering procedures is to find \emph{some} meaningful structure of the given data, and is not committed to any fixed objective function or any fixed number of clusters. The currently available theoretical research does not provide satisfactory formalizations of such ``flexible" clustering tasks\footnote{There haas been some recent work theoretically analyzing a notion of statistical stability (with respect to independent samplings) as a tool for determining an appropriate number of clusters as a function of the input data, e.g., \cite{BDPS07}, \cite{ShamirT10}. However, the conclusions if this work are mainly negative, showing that some proposed approaches may not work as intended.}, let alone an analysis of their computational complexity.
It may well be the case that our intuition of \emph{clustering being feasible on practically relevant cases} stems from clustering tasks that do not fit into the rigid fixed-$k$-fixed-objective framework of clustering.\\

\section*{Acknowledgements} I am grateful to Shalev Ben-David,  Lev Rayzin  and Ruth Urner  for insightful discussions concerning this paper. \\

%%%%%\bibliographystyle{plain}
%%%%%\bibliography{Researchbib}

\bibliographystyle{unsrt}
\bibliography{Union_clean_bibs}

\begin{thebibliography}{10}

\bibitem{DLS}
Amit Daniely, Nati Linial, and Michael Saks.
\newblock Clustering is difficult only when it does not matter.
\newblock {\em CoRR}, abs/1205.4891, 2012.

\bibitem{ben2015computational}
Shai Ben-David.
\newblock Computational feasibility of clustering under clusterability
  assumptions.
\newblock {\em arXiv preprint arXiv:1501.00437}, 2015.

\bibitem{AckermanB09}
Margareta Ackerman and Shai Ben-David.
\newblock Clusterability: A theoretical study.
\newblock In {\em AISTATS}, pages 1--8, 2009.

\bibitem{Ben-David06}
Shai Ben-David.
\newblock Alternative measures of computational complexity with applications to
  agnostic learning.
\newblock In {\em TAMC}, pages 231--235, 2006.

\bibitem{BiluL10}
Yonatan Bilu and Nathan Linial.
\newblock Are stable instances easy?
\newblock In {\em ICS}, pages 332--341, 2010.

\bibitem{OstrovskyRSS12}
Rafail Ostrovsky, Yuval Rabani, Leonard~J. Schulman, and Chaitanya Swamy.
\newblock The effectiveness of lloyd-type methods for the k-means problem.
\newblock {\em J. ACM}, 59(6):28, 2012.

\bibitem{AwasthiBS10}
Pranjal Awasthi, Avrim Blum, and Or~Sheffet.
\newblock Stability yields a ptas for k-median and k-means clustering.
\newblock In {\em FOCS}, pages 309--318, 2010.

\bibitem{AwasthiBS12}
Pranjal Awasthi, Avrim Blum, and Or~Sheffet.
\newblock Center-based clustering under perturbation stability.
\newblock {\em Inf. Process. Lett.}, 112(1-2):49--54, 2012.

\bibitem{BalcanBG13}
Maria-Florina Balcan, Avrim Blum, and Anupam Gupta.
\newblock Clustering under approximation stability.
\newblock {\em J. ACM}, 60(2):8, 2013.

\bibitem{BalcanBG09}
Maria-Florina Balcan, Avrim Blum, and Anupam Gupta.
\newblock Approximate clustering without the approximation.
\newblock In {\em SODA}, pages 1068--1077, 2009.

\bibitem{AwasthiBC14}
Pranjal Awasthi, Afonso Bandera, Moses Charikar, Ravishankar Krishnaswami,
  Soledad Voilar, and Rachel Ward.
\newblock Relax, no need to round: Integrality of clustering formulations.
\newblock {\em CoRR}, Stat.ML, 2014.

\bibitem{ShalevReyzin14}
Shalev Ben-David and Lev Reyzin.
\newblock Data stability in clustering: A closer look.
\newblock {\em Theoretical Computer Science}, page To appear, 2014.

\bibitem{Reyzin12}
Lev Reyzin.
\newblock Data stability in clustering: A closer look.
\newblock In {\em ALT}, pages 184--198, 2012.

\bibitem{BalcanL12}
Maria-Florina Balcan and Yingyu Liang.
\newblock Clustering under perturbation resilience.
\newblock In {\em ICALP (1)}, pages 63--74, 2012.

\bibitem{AckermanBL10}
Margareta Ackerman, Shai Ben{-}David, and David Loker.
\newblock Characterization of linkage-based clustering.
\newblock In {\em {COLT} 2010 - The 23rd Conference on Learning Theory, Haifa,
  Israel, June 27-29, 2010}, pages 270--281, 2010.

\end{thebibliography}

\end{document}